# A QUANTITATIVE ANALYSIS OF GRAPH REPRESENTATION STRATEGIES FOR CYBER-ATTACK DETECTION

**Ph.D., Ali Melih KANCA[1]**

[1]Department of Computer Engineering, Faculty of Computing and Informatics Sciences, Karabuk University, Karabuk, Türkiye

ORCID ID: https://orcid.org/0000-0003-3271-7463

melihkanca@karabuk.edu.tr +90 505 823 60 22

**Prof. İlker TÜRKER[2]**

[2]Computer Engineering, Faculty of Computing and Informatics Sciences, Karabuk University, Karabuk, Türkiye

ORCID ID: https://orcid.org/0000-0001-7577-4658

iturker@karabuk.edu.tr +90 370 418 92 22

## ABSTRACT

Graph-based cyber-attack detection studies employ various graph construction and representation strategies across different cybersecurity application domains. This diversity motivates a quantitative examination of how representation strategies are distributed across these application domains. This study presents a quantitative analysis of 37 original studies published between 2019 and 2026. Each study was coded according to publication year, application domain, graph representation type, feature extraction strategy, learning paradigm, algorithm, and dataset. Frequency analysis, cross-tabulation, and statistical association tests were applied. Automated representation learning was the most frequently employed strategy, accounting for 73.0% of the studies, while handcrafted representation strategies accounted for 27.0%. The Fisher–Freeman–Halton exact test revealed a statistically significant association between application domain and representation strategy (exact $p = 0.008$; Cramér's $V = 0.540$). The findings indicate that automated representation learning predominates within the analyzed corpus and that the distribution of representation strategies differs across cybersecurity application domains.

## INTRODUCTION

Graph-based approaches have been applied to a broad range of cyber-attack detection problems using different data modeling and learning strategies. Depending on the target application, network traffic, system interactions, and malicious behaviors can be represented through different graph structures, including communication graphs, visibility graphs, function call graphs, and provenance graphs. These representations can subsequently be analyzed using graph-theoretic features or automated representation-learning approaches (Lagraa et al., 2024; Zhong et al., 2024).

This diversity is reflected in the methodological choices reported across graph-based cyber-attack detection studies. While some approaches extract handcrafted topological features such as degree, centrality, and clustering coefficient and employ conventional machine learning classifiers (Shinan et al., 2023), recent studies increasingly utilize deep learning and Graph Neural Network (GNN) architectures to learn representations directly from graph structures. Moreover, the preferred graph representation, learning paradigm, and dataset vary considerably across application domains, including IDS/NIDS, botnet and DDoS detection, IoT-related threats, and advanced threats such as Android malware, Advanced Persistent Threats (APTs), and encrypted traffic (Caville et al., 2022; Farrukh et al., 2025; J. Liu & Guo, 2025; W. Liu et al., 2025; Pramanick et al., 2026; Zhao et al., 2026).

Motivated by this methodological diversity, this study presents a quantitative analysis of 37 original studies published between 2019 and 2026 on graph-based cyber-attack detection. The studies are systematically coded according to publication year, application domain, graph representation type, feature extraction strategy, learning paradigm, algorithm, and dataset. Frequency distributions, cross-tabulation, and statistical association tests are employed to investigate the relationships among these dimensions. The study particularly examines the distribution of handcrafted and automated representation strategies and their association with cybersecurity application domains.

Building upon the systematic review of graph-based representation techniques for cyber-attack detection presented in (Kanca & Türker, 2025), the present study incorporates recent studies published in 2025–2026 and focuses on a quantitative statistical analysis of representation strategies across application domains.

Unlike the systematic review in (Kanca & Türker, 2025), which primarily provided a qualitative synthesis and domain-based taxonomy, the present study quantitatively examines the association between application domain and representation strategy using cross-tabulation, the Fisher–Freeman–Halton exact test, and Cramér’s V.

## METHODOLOGY

### A. Study Corpus and Data Coding

The study corpus consists of 37 original studies published between 2019 and 2026 on graph-based cyber-attack detection. The studies cover four cybersecurity application domains: IDS/NIDS, Botnet/DDoS, IoT/DDoS, and Advanced Threats. Each study was treated as an individual observation and systematically coded to enable quantitative comparison across application domains and representation strategies.

The literature search was conducted across five major academic databases: IEEE Xplore, SpringerLink, ScienceDirect, Scopus, and Web of Science. The search strategy combined cybersecurity-related terms, including “cyberattack,” “intrusion detection,” “botnet detection,” “malware detection,” and “DDoS detection,” with graph-related terms such as “graph-based methods,” “graph representation,” “graph neural networks”, and “visibility graph”. These combinations were used to identify studies addressing graph-based representation and learning approaches for cyber-attack detection across different cybersecurity application domains.

The study corpus was derived from the systematic review framework presented in (Kanca & Türker, 2025) and was updated to incorporate relevant studies published in 2025–2026. Studies were included if they presented an original graph-based approach for cyber-attack detection and provided sufficient methodological information to identify the application domain, graph representation, representation strategy, and learning paradigm. Review articles, non-cybersecurity studies, studies not employing graph-based representations as a substantive component of the detection process, and studies lacking sufficient methodological information for the coding variables were excluded. Following these criteria and the update process, 37 original studies constituted the final corpus used for quantitative analysis.

The coding scheme included publication year, application domain, graph representation type, feature extraction strategy, learning paradigm, algorithm, dataset, and reported performance. Representation strategies were categorized as handcrafted or automated, while learning approaches were separately grouped into conventional machine learning, deep learning, GNN-based, and hybrid methods. This structured coding transformed the methodological characteristics of the selected studies into a dataset suitable for quantitative analysis.

Handcrafted strategies refer to approaches in which predefined graph-based or structural descriptors are explicitly extracted and used as representations, whereas automated strategies refer to approaches in which representations are learned or algorithmically derived from graph structures or attributes through methods such as graph embedding, deep learning, or graph neural networks.

The studies were classified into four cybersecurity application domains: IDS/NIDS (Alwasel et al., 2023; Caville et al., 2022; Chang & Branco, 2021; Desnitsky et al., 2023; Farrukh et al., 2025; Leichtnam et al., 2020; J. Liu & Guo, 2025; W. Liu et al., 2025; Pramanick et al., 2026; Sathishkumar et al., 2025; Su et al., 2019; Y. Wang et al., 2025; Yao et al., 2019; Zhao et al., 2026; Zhu & Lu, 2022; Zola et al., 2022), Botnet/DDoS (Alharbi & Alsubhi, 2021; Chatterjee, 2026; Daya et al., 2019; Lo et al., 2022; Millar et al., 2021; Shinan et al., 2023; Sinha et al., 2019; W. Wang et al., 2020), IoT/DDoS (Anjum et al., 2025; Jing & Wang, 2022; Khalaf et al., 2022; Nguyen et al., 2020; Protogerou et al., 2021; C. Wang & Zhu, 2022), and Advanced Threats (Cuzzocrea et al., 2023; Fu et al., 2022; Karrar, 2022; Li et al., 2021; H. Liu & Jiang, 2023; Niu et al., 2020; Wu et al., 2023). The variables and corresponding coding categories are summarized in Table 1.

**Table 1.** Study Coding Scheme.

| Variable | Coding / Description |
|---|---|
| Year | Publication year |
| Domain | IDS/NIDS, Botnet/DDoS, IoT/DDoS, Advanced Threats |
| Graph Type | Communication, VG/NVG/HVG, FCG, Provenance, Other |
| Representation Strategy | Handcrafted, Automated |
| Learning Paradigm | ML, DL, GNN, Hybrid |
| Algorithm | SVM, RF, CNN, LSTM, GCN, GAT, GraphSAGE, etc. |
| Dataset | Dataset(s) used in each study |
| Performance | Reported Accuracy/F1 when available |

The individual studies included in the quantitative analysis and their corresponding coding according to application domain, graph representation, representation strategy, and learning paradigm are presented in Table 2.

**Table 2.** Characteristics and Coding of the Studies Included in the Quantitative Analysis.

| Study ID | Source | Year | Application Domain | Graph Representation | Representation Strategy | Learning Paradigm |
|---|---|---|---|---|---|---|
| 1 | Alwasel et al. | 2023 | IDS/NIDS | Communication / neighborhood graph | Handcrafted | Classical ML / hybrid |
| 2 | Chang & Branco | 2021 | IDS/NIDS | Bipartite and line graph | Automated | GNN |
| 3 | Zhu & Lu | 2022 | IDS/NIDS | Line graph | Automated | GNN |
| 4 | Yao et al. | 2019 | IDS/NIDS | Temporal communication graph / graph kernel | Automated | DL |
| 5 | Su et al. | 2019 | IDS/NIDS | Graph kernel representation | Automated | DL |
| 6 | Leichtnam et al. | 2020 | IDS/NIDS | Security Objects Graph | Automated | Unsupervised DL / Autoencoder |
| 7 | Zola et al. | 2022 | IDS/NIDS | Temporal graph | Automated | GNN / DL hybrid |
| 8 | Desnitsky et al. | 2023 | IDS/NIDS | State-transition graph | Automated | Clustering / graph-based detection |
| 9 | Caville et al. | 2022 | IDS/NIDS | Edge-attributed graph | Automated | GNN |
| 10 | Wang et al. | 2020 | Botnet/DDoS | Communication / behavior graph | Handcrafted | Rule / similarity based |
| 11 | Millar et al. | 2021 | Botnet/DDoS | Flow communication graph | Automated | Graph embedding + ML |
| 12 | Shinan et al. | 2023 | Botnet/DDoS | Communication graph | Handcrafted | ML / DL comparison |
| 13 | Sinha et al. | 2019 | Botnet/DDoS | Communication graph | Handcrafted | DL |
| 14 | Alharbi & Alsubhi | 2021 | Botnet/DDoS | Communication graph | Handcrafted | Classical ML |
| 15 | Daya et al. | 2019 | Botnet/DDoS | Communication graph | Handcrafted | Unsupervised + Supervised ML |
| 16 | Lo et al. | 2022 | Botnet/DDoS | Flow graph | Automated | GNN |
| 17 | Khalaf et al. | 2022 | IoT/DDoS | Communication graph | Handcrafted | Classical ML |
| 18 | Protogerou et al. | 2021 | IoT/DDoS | Localized communication graph | Automated | GNN |
| 19 | Wang & Zhu | 2022 | IoT/DDoS | Behavioral Identification Graph | Automated | Graph embedding |
| 20 | Jing & Wang | 2022 | IoT/DDoS | Traffic / bot interaction graph | Handcrafted | Unsupervised ML |
| 21 | Nguyen et al. | 2020 | IoT/DDoS | PSI-rooted subgraph | Handcrafted | Classical ML |
| 22 | Niu et al. | 2020 | Advanced Threats | Opcode-level function call graph | Automated | DL |
| 23 | Wu et al. | 2023 | Advanced Threats | Static call / abstract flow graph | Automated | GNN / DL hybrid |
| 24 | Cuzzocrea et al. | 2023 | Advanced Threats | Control flow graph | Automated | Graph representation + ML |

| Study ID | Source | Year | Application Domain | Graph Representation | Representation Strategy | Learning Paradigm |
|---|---|---|---|---|---|---|
| 25 | Karrar | 2022 | Advanced Threats | Permission / API relation graph | Handcrafted | Classical ML |
| 26 | Li et al. | 2021 | Advanced Threats | Provenance and interhost graph | Automated | GNN |
| 27 | Fu et al. | 2022 | Advanced Threats | Heterogeneous spatio-temporal graph | Automated | Graph embedding |
| 28 | Liu & Jiang | 2023 | Advanced Threats | Causal graph | Automated | DL |
| 29 | Wang et al. | 2025 | IDS/NIDS | Behavior-similarity / network traffic graph | Automated | GNN |
| 30 | Liu & Guo | 2025 | IDS/NIDS | Dynamic / line graph | Automated | GNN |
| 31 | Farrukh et al. | 2025 | IDS/NIDS | Heterogeneous flow / packet graph | Automated | GNN |
| 32 | Anjum et al. | 2025 | IoT/DDoS | Dynamic IoT communication graph | Automated | GNN / hybrid |
| 33 | Liu et al. | 2025 | IDS/NIDS | Weighted k-NN graph | Automated | GNN |
| 34 | Sathishkumar et al. | 2025 | IDS/NIDS | Graph representation of network data | Automated | GNN |
| 35 | Chatterjee | 2026 | Botnet/DDoS | Dynamic IP / flow graph | Automated | GNN |
| 36 | Zhao et al. | 2026 | IDS/NIDS | Time-evolving dynamic spatio-temporal graph sequence | Automated | GNN / DL |
| 37 | Pramanick et al. | 2026 | IDS/NIDS | Cosine-similarity anchor graph | Automated | GNN |

### B. Quantitative and Statistical Analysis

The coded studies were analyzed using descriptive and inferential statistical methods. First, frequency and percentage distributions were calculated to examine the number of studies by publication year, application domain, and feature extraction strategy. Cross-tabulation was then employed to evaluate the distribution of handcrafted and automated representation strategies across the four cybersecurity application domains.

To statistically examine the relationship between application domain and representation strategy, handcrafted and automated approaches were compared across the four cybersecurity application domains. Because several expected cell frequencies were below five, the Fisher–Freeman–Halton exact test was used to evaluate the association between the two categorical variables (Mehta & Patel, 1983). Cramér's V, derived from the Pearson chi-square statistic, was additionally calculated to quantify the strength of the association.

## RESULTS AND ANALYSIS

### A. Distribution of Studies and Representation Strategies

A total of 37 original studies published between 2019 and 2026 were included in the quantitative analysis. Automated representation learning was the most frequently employed strategy, accounting for 27 studies (73.0%), while handcrafted graph representations were used in 10 studies (27.0%). These results indicate that automated representation learning constitutes the most frequently observed representation strategy within the analyzed corpus. The overall distribution of representation strategies is presented in Table 3 and visually summarized in Figure 1.

**Table 3.** Distribution of Representation Strategies**.**

| Feature Strategy | Number of Studies | Percentage |
|---|---|---|
| Automated | 27 | 73% |
| Handcrafted | 10 | 27% |
| Total | 37 | 100.0% |

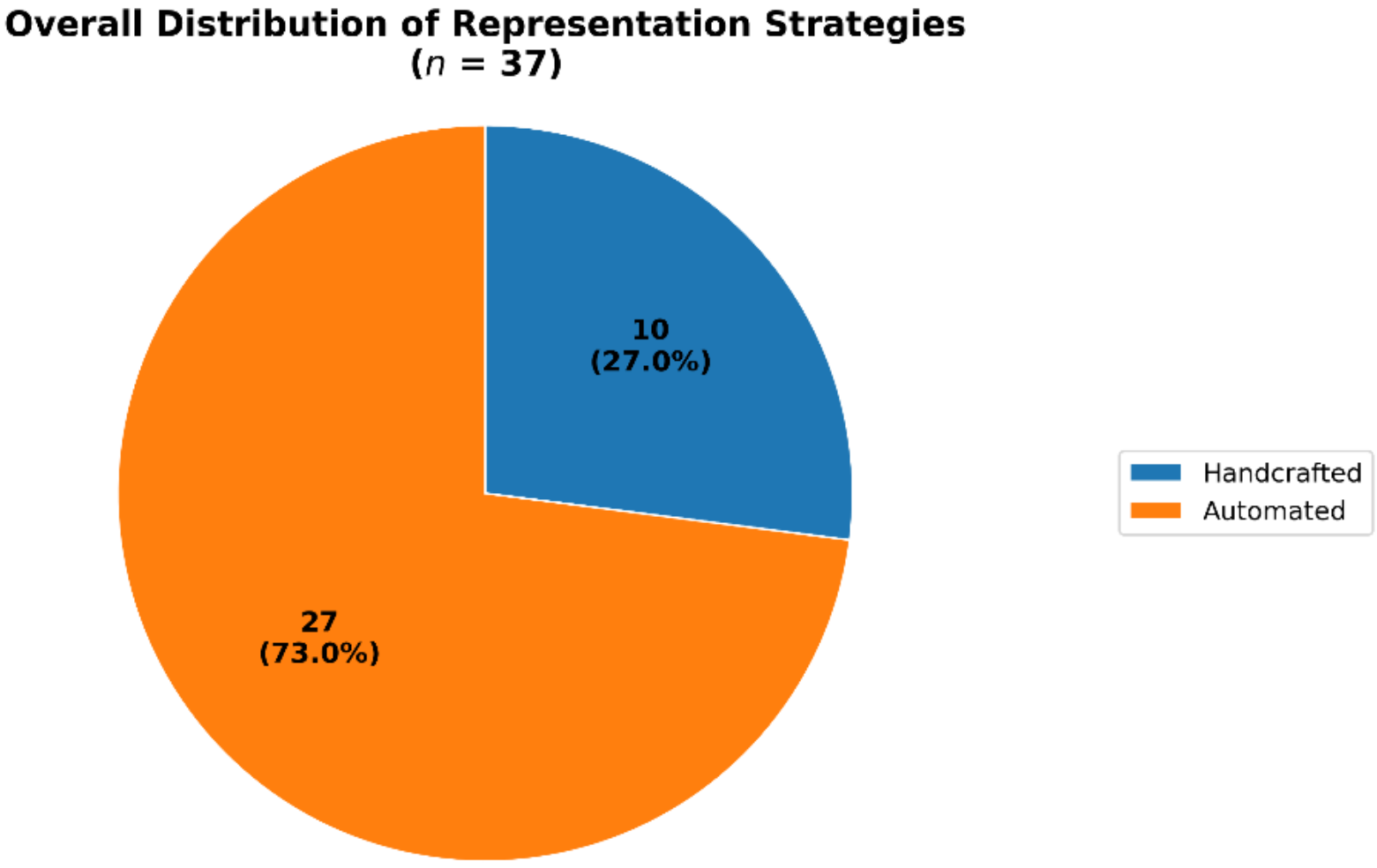

**Figure 1.** Overall distribution of representation strategies in the analyzed corpus (n = 37).

### B. Feature Strategy Across Application Domains

IDS/NIDS represented the largest application domain with 16 studies (43.2%), followed by Botnet/DDoS with eight studies (21.6%), Advanced Threats with seven studies (18.9%), and IoT/DDoS with six studies (16.2%). The distribution of representation strategies differed considerably across these application domains.

Automated approaches accounted for 15 of the 16 IDS/NIDS studies (93.8%) and six of the seven Advanced Threats studies (85.7%). In contrast, handcrafted graph features were used in five of the eight Botnet/DDoS studies (62.5%) and three of the six IoT/DDoS studies (50.0%). These distributions indicate a clear preference for automated representation learning in IDS/NIDS and Advanced Threats, whereas handcrafted representations remain more prevalent in Botnet/DDoS and are equally represented with automated approaches in IoT/DDoS applications. The domain-specific distribution of representation strategies is reported in Table 4 and illustrated in Figure 2.

**Table 4**. Representation Strategy by Application Domain**.**

| Application Domain | Handcrafted | Automated |
|---|---|---|
| IDS/NIDS | 1 | 15 |
| Botnet/DDoS | 5 | 3 |
| IoT/DDoS | 3 | 3 |
| Advanced Threats | 1 | 6 |
| Total | 10 | 27 |

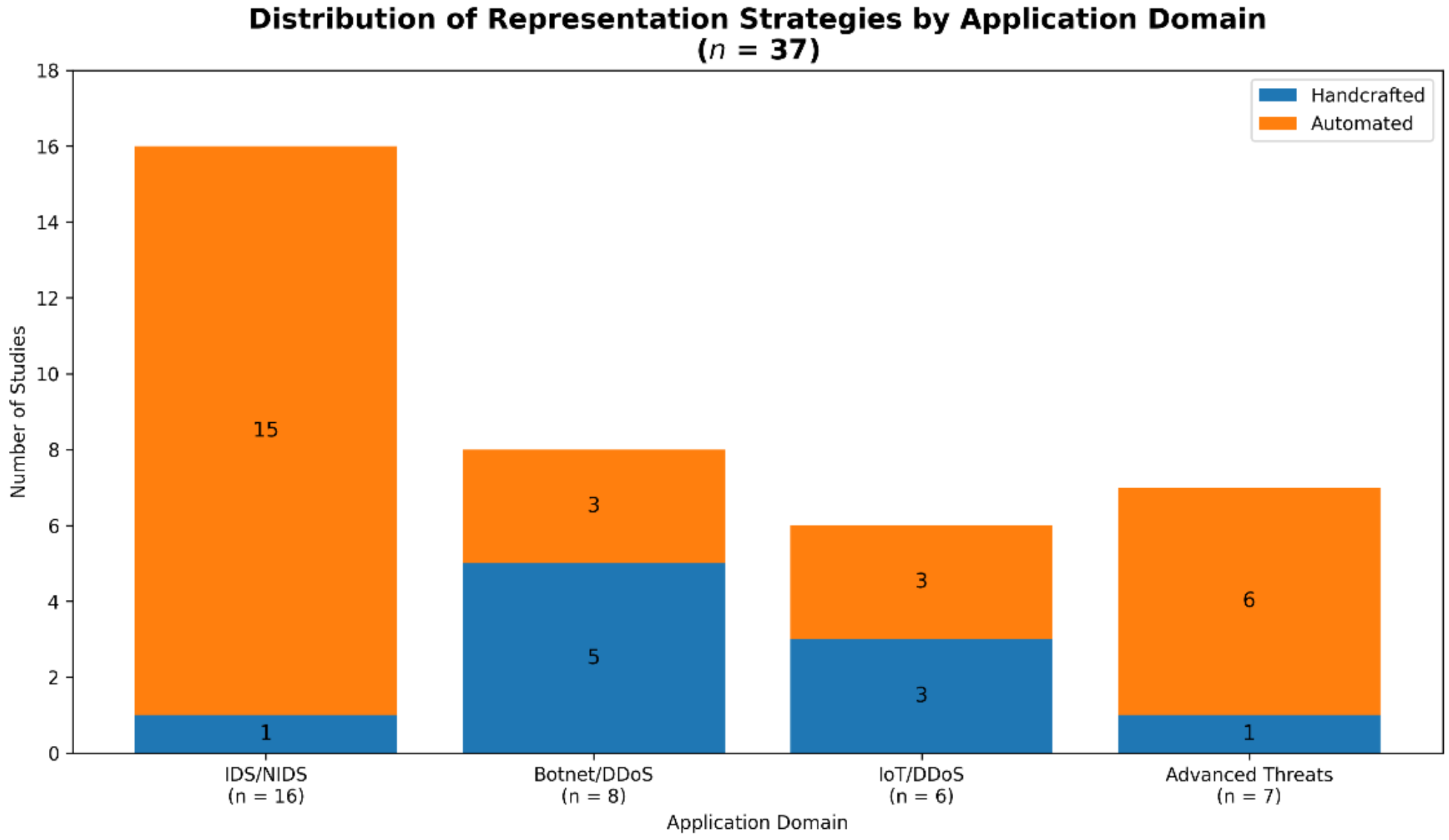

**Figure 2.** Distribution of representation strategies across cybersecurity application domains (n = 37).

Because several expected frequencies were below five, the association was evaluated using the Fisher–Freeman–Halton exact test. The analysis revealed a statistically significant association between application domain and representation strategy (exact $p = 0.008$). The corresponding Cramér's V was 0.540. These findings indicate a statistically significant association between application domain and representation strategy within the analyzed corpus.

## CONCLUSION

This study presented a quantitative analysis of graph-based cyber-attack detection research based on 37 original studies published between 2019 and 2026. Automated representation learning was the most frequently employed strategy, accounting for 73.0% of the studies, while handcrafted approaches represented 27.0%. The distribution of representation strategies varied across application domains. Automated approaches were more prevalent in IDS/NIDS and Advanced Threats, whereas handcrafted graph representations were more common in Botnet/DDoS. In IoT/DDoS applications, handcrafted and automated representation strategies were equally represented. The Fisher–Freeman–Halton exact test indicated a statistically significant association between application domain and representation strategy (exact $p = 0.008$), with a Cramér's V of 0.540.

The observed distribution suggests that representation strategies are not uniformly adopted across cybersecurity application domains. This pattern may reflect differences in the types of graph structures, data characteristics, and analytical objectives encountered across these domains. However, the present analysis establishes an association rather than a causal relationship, and further research is required to determine the methodological factors underlying these domain-specific patterns.

The study is limited by the size of the analyzed corpus and by methodological heterogeneity among the included studies. Dataset and performance values were not directly compared because the studies employed different datasets, preprocessing procedures, classification tasks, and evaluation metrics. Future research may extend the corpus and investigate dataset concentration, reporting practices, learning paradigms, and graph-representation choices using larger study samples.

**REFERENCES**

Alharbi, A., & Alsubhi, K. (2021). Botnet Detection Approach Using Graph-Based Machine Learning. *IEEE Access*, *9*, 99166–99180. https://doi.org/10.1109/ACCESS.2021.3094183

Alwasel, B., Aldribi, A., Alreshoodi, M., Alsukayti, I. S., & Alsuhaibani, M. (2023). Leveraging Graph-Based Representations to Enhance Machine Learning Performance in IIoT Network Security and Attack Detection. *Applied Sciences (Switzerland)*, *13*(13). https://doi.org/10.3390/app13137774

Anjum, M., Dutta, A. K., Elrashidi, A., Shahab, S., Aldrees, A., Shaikh, Z. A., & Aljohani, A. (2025). GraphFedAI framework for DDoS attack detection in IoT systems using federated learning and graph based artificial intelligence. *Scientific Reports 2025 15:1*, *15*(1), 28050-. https://doi.org/10.1038/s41598-025-10826-0

Caville, E., Lo, W. W., Layeghy, S., & Portmann, M. (2022). Anomal-E: A self-supervised network intrusion detection system based on graph neural networks. *Knowledge-Based Systems*, *258*. https://doi.org/10.1016/j.knosys.2022.110030

Chang, L., & Branco, P. (2021). *Graph-based Solutions with Residuals for Intrusion Detection: the Modified E-GraphSAGE and E-ResGAT Algorithms*. http://arxiv.org/abs/2111.13597

Chatterjee, S. (2026). Robust DDoS detection based on adversarial trained graph neural networks with proximal optimization. *Social Network Analysis and Mining 2026 16:1*, *16*(1), 39-. https://doi.org/10.1007/S13278-026-01578-4

Cuzzocrea, A., Quebrado, M., Hafsaoui, A., & Serra, E. (2023). A Graph-Representation-Learning Framework for Supporting Android Malware Identification and Polymorphic Evolution. *Proceedings - 2023 10th IEEE Swiss Conference on Data Science, SDS 2023*, 34–41. https://doi.org/10.1109/SDS57534.2023.00012

Daya, A. A., Salahuddin, M. A., Limam, N., & Boutaba, R. (2019). A Graph-Based Machine Learning Approach for Bot Detection. *2019 IFIP/IEEE Symposium on Integrated*

*Network and Service Management, IM 2019*, 144–152. https://arxiv.org/abs/1902.08538v1

Desnitsky, V., Levshun, D., & Kotenko, I. (2023). An Approach to Attack Detection on Graph-Based Modeling of States in Critical Infrastructures. *Proceedings - 2023 International Russian Automation Conference, RusAutoCon 2023*, 420–424. https://doi.org/10.1109/RusAutoCon58002.2023.10272943

Farrukh, Y. A., Wali, S., Khan, I., & Bastian, N. D. (2025). XG-NID: Dual-modality network intrusion detection using a heterogeneous graph neural network and large language model. *Expert Systems with Applications*, *287*, 128089. https://doi.org/10.1016/J.ESWA.2025.128089

Fu, Z., Liu, M., Qin, Y., Zhang, J., Zou, Y., Yin, Q., Li, Q., & Duan, H. (2022). Encrypted Malware Traffic Detection via Graph-based Network Analysis. *ACM International Conference Proceeding Series*, 495–509. https://doi.org/10.1145/3545948.3545983

Jing, H., & Wang, J. (2022). DDoS detection based on graph structure features and non-negative matrix factorization. *Concurrency and Computation: Practice and Experience*, *34*(9). https://doi.org/10.1002/cpe.5783

Kanca, A. M., & Türker, İ. (2025). A Systematic Review of Graph-Based Representation Techniques for Cyber-Attack Detection Across Application Domains. *Concurrency and Computation: Practice and Experience*, *37*(27–28). https://doi.org/10.1002/CPE.70389

Karrar, A. E. (2022). Adopting Graph-Based Machine Learning Algorithms to Classify Android Malware. *IJCSNS International Journal of Computer Science and Network Security*, *22*(9), 840. https://doi.org/10.22937/IJCSNS.2022.22.9.109

Khalaf, O. I., Ogudo, K. A., & Sangeetha, S. K. B. (2022). Design of Graph-Based Layered Learning-Driven Model for Anomaly Detection in Distributed Cloud IoT Network. *Mobile Information Systems*, *2022*. https://doi.org/10.1155/2022/6750757

Lagraa, S., Husák, M., Seba, H., Vuppala, S., State, R., & Ouedraogo, M. (2024). A review on graph-based approaches for network security monitoring and botnet detection. *International Journal of Information Security*, *23*(1), 119–140. https://doi.org/10.1007/S10207-023-00742-7

Leichtnam, L., Totel, E., Prigent, N., & Mé, L. (2020). Sec2graph: Network Attack Detection Based on Novelty Detection on Graph Structured Data. *Lecture Notes in Computer Science (Including Subseries Lecture Notes in Artificial Intelligence and Lecture Notes in Bioinformatics)*, *12223 LNCS*, 238–258. https://doi.org/10.1007/978-3-030-52683-2_12

Li, Z., Cheng, X., Sun, L., Zhang, J., & Chen, B. (2021). A Hierarchical Approach for Advanced Persistent Threat Detection with Attention-Based Graph Neural Networks. *Security and Communication Networks*, *2021*. https://doi.org/10.1155/2021/9961342

Liu, H., & Jiang, R. (2023). A Causal Graph-Based Approach for APT Predictive Analytics. *Electronics (Switzerland)*, *12*(8). https://doi.org/10.3390/electronics12081849

Liu, J., & Guo, M. (2025). DIGNN-A: Real-Time Network Intrusion Detection with Integrated Neural Networks Based on Dynamic Graph. *Computers, Materials and Continua*, *82*(1), 817–842. https://doi.org/10.32604/CMC.2024.057660

Liu, W., Li, S., Gao, C., Sang, Y., & Ge, H. (2025). An adaptive graph neural network-based intrusion detection system for airborne network. *Engineering Applications of Artificial Intelligence*, *152*, 110851. https://doi.org/10.1016/J.ENGAPPAI.2025.110851

Lo, W. W., Layeghy, S., Sarhan, M., Gallagher, M., & Portmann, M. (2022). E-GraphSAGE: A Graph Neural Network based Intrusion Detection System for IoT. *Proceedings of the IEEE/IFIP Network Operations and Management Symposium 2022: Network and Service Management in the Era of Cloudification, Softwarization and Artificial Intelligence, NOMS 2022*. https://doi.org/10.1109/NOMS54207.2022.9789878

Mehta, C. R., & Patel, N. R. (1983). A Network Algorithm for Performing Fisher's Exact Test in r × c Contingency Tables. *Journal of the American Statistical Association*, *78*(382), 427–434. https://doi.org/10.1080/01621459.1983.10477989

Millar, K., Simpson, L., Cheng, A., Chew, H. G., & Lim, C. C. (2021). DETECTING BOTNET VICTIMS THROUGH GRAPH-BASED MACHINE LEARNING. *Proceedings - International Conference on Machine Learning and Cybernetics*, *2021-December*. https://doi.org/10.1109/ICMLC54886.2021.9737249

Nguyen, H. T., Ngo, Q. D., Nguyen, D. H., & Le, V. H. (2020). PSI-rooted subgraph: A novel feature for IoT botnet detection using classifier algorithms. *ICT Express*, *6*(2), 128–138. https://doi.org/10.1016/J.ICTE.2019.12.001

Niu, W., Cao, R., Zhang, X., Ding, K., Zhang, K., & Li, T. (2020). Opcode-level function call graph based android malware classification using deep learning. *Sensors (Switzerland)*, *20*(13), 1–23. https://doi.org/10.3390/s20133645

Pramanick, N., Mathew, J., Nam, S. Y., & Agarwal, M. (2026). ADA-GNetIDS: ADASYN-augmented graph neural network based learning in IDS for robust cyber attack classification. *Scientific Reports 2026*. https://doi.org/10.1038/S41598-026-69410-9

Protogerou, A., Papadopoulos, S., Drosou, A., Tzovaras, D., & Refanidis, I. (2021). A graph neural network method for distributed anomaly detection in IoT. *Evolving Systems*, *12*(1), 19–36. https://doi.org/10.1007/s12530-020-09347-0

Sathishkumar, P., Nikitha, S., Sruthi, R., & Vishwa, R. R. (2025). *Anomaly Detection in Network Security Using Unsupervised Graph Neural Network*. 1–7. https://doi.org/10.1109/ICOACT63339.2025.11004728

Shinan, K., Alsubhi, K., & Usman Ashraf, M. (2023). BotSward: Centrality Measures for Graph-Based Bot Detection Using Machine Learning. *Computers, Materials and Continua*, *74*(1), 693–714. https://doi.org/10.32604/cmc.2023.031641

Sinha, K., Viswanathan, A., & Bunn, J. (2019). *Tracking Temporal Evolution of Network Activity for Botnet Detection*. https://arxiv.org/abs/1908.03443v1

Su, L., Yao, Y., Lu, Z., & Liu, B. (2019). Understanding the influence of graph kernels on deep learning architecture: A case study of flow-based network attack detection. *Proceedings - 2019 18th IEEE International Conference on Trust, Security and Privacy in Computing and Communications/13th IEEE International Conference on Big Data Science and Engineering, TrustCom/BigDataSE 2019*, 312–318. https://doi.org/10.1109/TrustCom/BigDataSE.2019.00049

Wang, C., & Zhu, H. (2022). Wrongdoing Monitor: A Graph-Based Behavioral Anomaly Detection in Cyber Security. *IEEE Transactions on Information Forensics and Security*, *17*, 2703–2718. https://doi.org/10.1109/TIFS.2022.3191493

Wang, W., Shang, Y., He, Y., Li, Y., & Liu, J. (2020). BotMark: Automated botnet detection with hybrid analysis of flow-based and graph-based traffic behaviors. *Information Sciences*, *511*, 284–296. https://doi.org/10.1016/j.ins.2019.09.024

Wang, Y., Han, Z., Du, Y., Li, J., & He, X. (2025). BS-GAT: a network intrusion detection system based on graph neural network for edge computing. *Cybersecurity 2025 8:1*, *8*(1), 27-. https://doi.org/10.1186/S42400-024-00296-8

Wu, Y., Shi, J., Wang, P., Zeng, D., & Sun, C. (2023). DeepCatra: Learning flow- and graph-based behaviours for Android malware detection. *IET Information Security*, *17*(1), 118–130. https://doi.org/10.1049/ise2.12082

Yao, Y., Su, L., Lu, Z., & Liu, B. (2019). STDeepGraph: Spatial-temporal deep learning on communication graphs for long-term network attack detection. *Proceedings - 2019 18th IEEE International Conference on Trust, Security and Privacy in Computing and Communications/13th IEEE International Conference on Big Data Science and Engineering, TrustCom/BigDataSE 2019*, 120–127. https://doi.org/10.1109/TrustCom/BigDataSE.2019.00025

Zhao, J., Zhang, D., He, Q., Lin, M., & Yang, Y. (2026). DMSTG-AD: an SDN intrusion detection method based on dynamic multi-scale spatio-temporal graph neural network. *Scientific Reports 2026 16:1*, *16*(1), 14528-. https://doi.org/10.1038/s41598-026-44360-4

Zhong, M., Lin, M., Zhang, C., & Xu, Z. (2024). A survey on graph neural networks for intrusion detection systems: Methods, trends and challenges. *Computers & Security*, *141*, 103821. https://doi.org/10.1016/J.COSE.2024.103821

Zhu, H., & Lu, J. (2022). Graph-based Intrusion Detection System Using General Behavior Learning. *2022 IEEE Global Communications Conference, GLOBECOM 2022 - Proceedings*, 2621–2626. https://doi.org/10.1109/GLOBECOM48099.2022.10001299

Zola, F., Segurola-Gil, L., Bruse, J. L., Galar, M., & Orduna-Urrutia, R. (2022). Network traffic analysis through node behaviour classification: a graph-based approach with temporal dissection and data-level preprocessing. *Computers and Security*, *115*. https://doi.org/10.1016/j.cose.2022.102632